\documentclass{article}

\usepackage{microtype}
\usepackage{graphicx}
\usepackage{subfigure}
\usepackage{booktabs} % for professional tables

\usepackage{hyperref}

\usepackage[accepted]{icml2024}

\usepackage{amsmath}
\usepackage{amssymb}
\usepackage{mathtools}
\usepackage{amsthm}

\usepackage[capitalize,noabbrev]{cleveref}

\theoremstyle{plain}

\theoremstyle{definition}

\theoremstyle{remark}

\newcommand*{\diris}{$\Delta$-\textsc{iris}}
\newcommand*{\dtoken}{$\Delta$}
\newcommand*{\itoken}{\scalebox{0.9}[0.9]{I}}

\usepackage{tikz}
\usetikzlibrary{arrows,arrows.meta,calc}
\usetikzlibrary{patterns,backgrounds}
\usetikzlibrary{positioning,fit}
\usetikzlibrary{shapes.geometric,shapes.multipart}
\usetikzlibrary{patterns.meta,decorations.pathreplacing,calligraphy}
\usetikzlibrary{tikzmark}
\usetikzlibrary{decorations.pathmorphing}

\tikzset{
  every node/.style={font=\tiny,inner sep=0pt},
  font=\small,
  empty/.style={minimum width=6.7mm,minimum height=0.65cm},
  ztoken/.style={empty,draw},
  action/.style={empty,draw},
  imgstd/.style={empty, draw},
  img/.style={imgstd,anchor=center},
  z small/.style={minimum width=1cm, minimum height=0.65cm,draw},
  z large/.style={minimum width=2cm, minimum height=0.65cm,draw},
  next1/.style={right=1.5cm of \tikzlastnode},
  next2/.style={right=3cm of \tikzlastnode},
  next/.style={right=2pt of \tikzlastnode},
  flow/.style={->,thick,shorten >=2pt,shorten <=2pt,},
  flow headless/.style={thick,shorten >=2pt,shorten <=2pt,},
  computed/.style={fill=red!35},
  true/.style={fill=blue!35},
  curly brace/.style={sharp corners,decoration={brace,amplitude=0.15cm},decorate},
}
\usepackage[textsize=tiny]{todonotes}

\icmltitlerunning{Efficient World Models with Context-Aware Tokenization}

\begin{document}

\twocolumn[
\icmltitle{Efficient World Models with Context-Aware Tokenization}



\icmlsetsymbol{equal}{*}

\begin{icmlauthorlist}
\icmlauthor{Vincent Micheli}{equal,unige}
\icmlauthor{Eloi Alonso}{equal,unige}
\icmlauthor{François Fleuret}{unige}
\end{icmlauthorlist}

\icmlaffiliation{unige}{University of Geneva, Switzerland}

\icmlcorrespondingauthor{}{first.last@unige.ch}

\icmlkeywords{Artificial Intelligence, Machine Learning, Generative Modelling, World Models, Transformers, Autoencoders, Reinforcement Learning, Decision making, ICML}

\vskip 0.3in
]



\printAffiliationsAndNotice{\icmlEqualContribution} % otherwise use the standard text.

\begin{abstract}
Scaling up deep Reinforcement Learning (RL) methods presents a significant challenge. Following developments in generative modelling, model-based RL positions itself as a strong contender.
Recent advances in sequence modelling have led to effective transformer-based world models, albeit at the price of heavy computations due to the long sequences of tokens required to accurately simulate environments.
In this work, we propose \diris, a new agent with a world model architecture composed of a discrete autoencoder that encodes stochastic deltas between time steps and an autoregressive transformer that predicts future deltas by summarizing the current state of the world with continuous tokens.
In the Crafter benchmark, \diris\ sets a new state of the art at multiple frame budgets, while being an order of magnitude faster to train than previous attention-based approaches. We release our code and models at {\small \url{https://github.com/vmicheli/delta-iris}}.
\end{abstract}

\section{Introduction}
\label{sec:introduction}

Deep Reinforcement Learning (RL) methods have recently delivered impressive results \cite{efficientzero, dreamerv3, schwarzer2023bigger} in traditional benchmarks \cite{bellemare13arcade, dmc}. In light of the evermore complex domains tackled by the latest generations of generative models \cite{stablediffusion, gpt4}, the prospect of training agents in more ambitious environments \cite{mineRL} may hold significant appeal. However, that leap forward poses a serious challenge: deep RL architectures have been comparatively smaller and less sample-efficient than their (self-)supervised counterparts. In contrast, more intricate environments necessitate models with greater representational power and have higher data requirements.

Model-based RL (MBRL) \cite{sutton} is hypothesized to be the key for scaling up deep RL agents \cite{lecun2022path}. Indeed, world models \cite{worldmodels} offer a diverse range of capabilities: lookahead search \cite{muzero, efficientzero}, learning in imagination \cite{sutton1991dyna, dreamerv3}, representation learning \cite{spr, srspr}, and uncertainty estimation \cite{curiosity, plan2explore}. In essence, MBRL shifts the focus from the RL problem to a generative modelling problem, where the development of an accurate world model significantly simplifies policy training. In particular, policies learnt in the imagination of world models are freed from sample efficiency constraints, a common limitation of RL agents that is magnified in complex environments with slow rollouts.

Recently, the \textsc{iris} agent \cite{iris} achieved strong results in the Atari 100k benchmark \cite{bellemare13arcade, simple}. \textsc{iris} introduced a world model composed of a discrete autoencoder and an autoregressive transformer, casting dynamics learning as a sequence modelling problem where the transformer composes over time a vocabulary of image tokens built by the autoencoder. This approach opened up avenues for future model-based methods to capitalize on advances in generative modelling \cite{phenaki, gpt4}, and has already been adopted beyond its original domain \cite{commavq, hu2023gaia}. However, in its current form, scaling \textsc{iris} to more complex environments is computationally prohibitive. Indeed, such an endeavor requires a large number of tokens to encode visually challenging frames. Besides, sophisticated dynamics may require to store numerous time steps in memory to reason about the past, ultimately making the imagination procedure excessively slow. Hence, under these constraints, maintaining a favorable imagined-to-collected data ratio is practically infeasible.

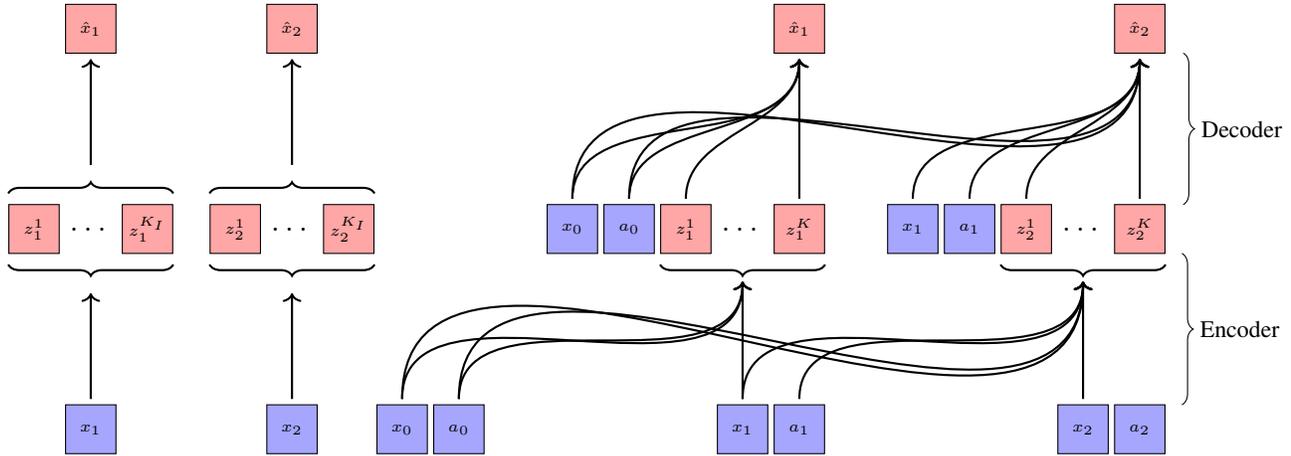
\begin{figure*}[t]
\center

\begin{tikzpicture}[font=\scriptsize, scale=0.95]

  \node[] at (0, 0) {%
    \begin{tikzpicture}
  \path
  node[computed,img] (enc z1 1) {$z^1_1$}
  node[empty,next] (enc zdots 1) {\large $\dots$}
  node[computed,img,next] (enc zK 1) {$z^{K_I}_1$}
  node[empty,next] (empty) {}
  node[computed,next,img] (enc z1 2) {$z^1_2$}
  node[empty,next] (enc zdots 2) {\large $\dots$}
  node[computed,img,next] (enc zK 2) {$z^{K_I}_2$}
  ;

  \node[true,img,below=2cm of enc zdots 1] (enc x1) {$x_1$};
  \node[computed,img,above=2cm of enc zdots 1] (enc hatx1) {$\hat{x}_1$};

  \draw[curly brace,thick] ($(enc zK 1.south east)+(0,-4pt)$) -- ($(enc z1 1.south west) + (0,-4pt)$) node[midway,below,inner sep=4pt] (enc zgroup down 1) {};
  \draw[curly brace,thick] ($(enc z1 1.north west)+(0,4pt)$) -- ($(enc zK 1.north east) + (0,4pt)$) node[midway,above,inner sep=4pt] (enc zgroup up 1) {};

  \draw[flow] (enc x1) to[out=90,in=270] (enc zgroup down 1);
  \draw[flow] (enc zgroup up 1) to[out=90,in=270] (enc hatx1);

  \node[true,img,below=2cm of enc zdots 2] (enc x2) {$x_2$};
  \node[computed,img,above=2cm of enc zdots 2] (enc hatx2) {$\hat{x}_2$};

  \draw[curly brace,thick] ($(enc zK 2.south east)+(0,-4pt)$) -- ($(enc z1 2.south west) + (0,-4pt)$) node[midway,below,inner sep=4pt] (enc zgroup down 2) {};
  \draw[curly brace,thick] ($(enc z1 2.north west)+(0,4pt)$) -- ($(enc zK 2.north east) + (0,4pt)$) node[midway,above,inner sep=4pt] (enc zgroup up 2) {};

  \draw[flow] (enc x2) to[out=90,in=270] (enc zgroup down 2);
  \draw[flow] (enc zgroup up 2) to[out=90,in=270] (enc hatx2);
    \end{tikzpicture}%
  };

\node[] at (9,0) {%
\begin{tikzpicture}[]

\path
node[true,img] (enc x0) {$x_0$}
node[true,img,next] (enc a0) {$a_0$}
node[computed,img,next] (enc z1 1) {$z^1_1$}
node[empty,next] (enc zdots 1) {\large $\dots$}
node[computed,img,next] (enc zK 1) {$z^K_1$}
node[empty,next] (enc empty) {}
node[true,img,next] (enc x1) {$x_1$}
node[true,img,next] (enc a1) {$a_1$}
node[computed,img,next] (enc z1 2) {$z^1_2$}
node[empty,next] (enc zdots 2) {\large $\dots$}
node[computed,img,next] (enc zK 2) {$z^K_2$}
;

\draw[curly brace,thick] ($(enc zK 1.south east)+(0,-4pt)$) -- ($(enc z1 1.south west) + (0,-4pt)$) node[midway,below,inner sep=4pt] (enc zgroup 1) {};
\draw[curly brace,thick] ($(enc zK 2.south east)+(0,-4pt)$) -- ($(enc z1 2.south west) + (0,-4pt)$) node[midway,below,inner sep=4pt] (enc zgroup 2) {};

\node[true,img,below=2cm of {enc zdots 1}] (x1) {$x_1$};
\node[true,img,below=2cm of {enc zdots 2}] (x2) {$x_2$};
\node[true,img] (x0) at ($(x1) + (x1) - (x2)$) {$x_0$};
\node[true,action,right=2pt of x0] (a0) {$a_0$};
\node[true,action,right=2pt of x1] (a1) {$a_1$};
\node[true,action,right=2pt of x2] (a2) {$a_2$};

\node[computed,img,above=2cm of enc zK 1] (enc hatx1) {$\hat{x}_1$};
\node[computed,img,above=2cm of enc zK 2] (enc hatx2) {$\hat{x}_2$};

\node[xshift=6pt,fit=(a0.north east) (enc zK 2.south east)] (encoder) {};
\draw[curly brace] (encoder.north east) -- (encoder.south east)
node[midway,right,xshift=1ex,inner sep=4pt,inner sep=4pt] {\small Encoder};

\node[xshift=6pt,fit=(enc hatx2.south east) (enc zK 2.north east)] (decoder) {};
\draw[curly brace] (decoder.north east) -- (decoder.south east)
node[midway,right,xshift=1ex,inner sep=4pt,inner sep=4pt] {\small Decoder};

%------------------------------------------------------

\begin{pgfinterruptboundingbox}

\draw[flow] (x0) to[out=90,in=270]  (enc zgroup 1.center);
\draw[flow] (x0) to[out=90,in=270]  (enc zgroup 2.center);
\draw[flow] (x1) to[out=90,in=270]  (enc zgroup 1.center);
\draw[flow] (x1) to[out=90,in=270]  (enc zgroup 2.center);
\draw[flow] (x2) to[out=90,in=270]  (enc zgroup 2.center);
\draw[flow] (a0) to[out=90,in=270]  (enc zgroup 1.center);
\draw[flow] (a0) to[out=90,in=270]  (enc zgroup 2.center);
\draw[flow] (a1) to[out=90,in=270]  (enc zgroup 2.center);

\draw[flow] (enc x0) to[out=90,in=270] (enc hatx1);
\draw[flow] (enc a0) to[out=90,in=270] (enc hatx1);
\draw[flow] (enc z1 1) to[out=90,in=270] (enc hatx1);
\draw[flow] (enc zK 1) to[out=90,in=270] (enc hatx1);

\draw[flow] (enc x0) to[out=90,in=270] (enc hatx2);
\draw[flow] (enc a0) to[out=90,in=270] (enc hatx2);
\draw[flow] (enc x1) to[out=90,in=270] (enc hatx2);
\draw[flow] (enc a1) to[out=90,in=270] (enc hatx2);
\draw[flow] (enc z1 2) to[out=90,in=270] (enc hatx2);
\draw[flow] (enc zK 2) to[out=90,in=270] (enc hatx2);

\end{pgfinterruptboundingbox}
%------------------------------------------------------

\end{tikzpicture}
};
\end{tikzpicture}

\caption{Discrete autoencoder of \textsc{iris} \cite{iris} (left) and \diris\ (right). \textsc{iris} encodes and decodes frames independently, meaning that $z_t$ has to carry all the information necessary to reconstruct $x_t$. On the other hand, \diris' encoder and decoder are conditioned on past frames and actions, thus $z_t$ only has to capture what has changed and that cannot be inferred from actions, i.e. the stochastic delta. This conditioning scheme enables us to drastically reduce the number of tokens required to encode a frame with minimal loss ($K \ll K_I$), which is critical to speed up the autoregressive transformer that predicts future tokens.}
\label{fig:autoencoder}
\end{figure*}

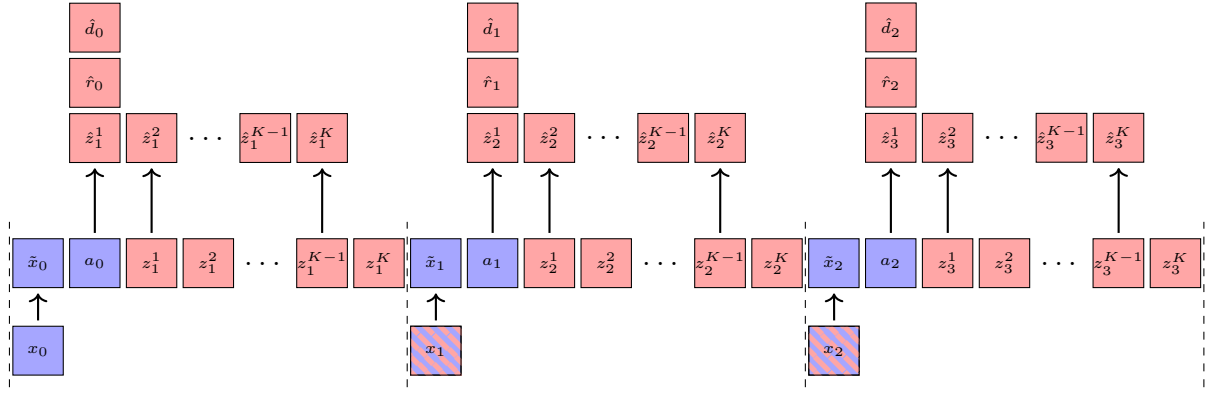
\begin{figure*}[t]
\center

\begin{tikzpicture}[font=\scriptsize, scale=1.1]

\path
node[empty] (phantom) {}
node[next,true,imgstd] (x0 tilde) {$\tilde{x}_0$}
node[next,true,action] (a0) {$a_0$}
node[next,ztoken,computed] (z1 1) {$z^1_1$}
node[next,ztoken,computed] (z2 1) {$z^2_1$}
node[next,empty] (z etc 1) {\large $\dots$}
node[next,ztoken,computed] (zKm1 1) {$z^{K-1}_1$}
node[next,ztoken,computed] (zK 1) {$z^K_1$}
node[next,true,imgstd] (x1 tilde) {$\tilde{x}_1$}
node[next,true,action] (a1) {$a_1$}
%% node[next,computed,z small,minimum width=1.8cm] (z2) {$z^1_2,\dots,z^K_2$}
node[next,ztoken,computed] (z1 2) {$z^1_2$}
node[next,ztoken,computed] (z2 2) {$z^2_2$}
node[next,empty] (z etc 2) {\large $\dots$}
node[next,ztoken,computed] (zKm1 2) {$z^{K-1}_2$}
node[next,ztoken,computed] (zK 2) {$z^K_2$}
node[next,true,imgstd] (x2 tilde) {$\tilde{x}_2$}
node[next,true,action] (a2) {$a_2$}
%% node[next,computed,z small,minimum width=1.8cm] (z2) {$z^1_2,\dots,z^K_2$}
node[next,ztoken,computed] (z1 3) {$z^1_3$}
node[next,ztoken,computed] (z2 3) {$z^2_3$}
node[next,empty] (z etc 3) {\large $\dots$}
node[next,ztoken,computed] (zKm1 3) {$z^{K-1}_3$}
node[next,ztoken,computed] (zK 3) {$z^K_3$}
node[next,empty] (x3 tilde) {}
;

\draw[dashed] ($(x0 tilde.west)+(-1pt,0)+(0,-1.5)$) -- ++(0,2);
\draw[dashed] ($(x1 tilde.west)+(-1pt,0)+(0,-1.5)$) -- ++(0,2);
\draw[dashed] ($(x2 tilde.west)+(-1pt,0)+(0,-1.5)$) -- ++(0,2);
\draw[dashed] ($(x3 tilde.west)+(-1pt,0)+(0,-1.5)$) -- ++(0,2);

\path
node[empty,above=1cm of x0 tilde] (phantom next) {}
%% node[above=2cm of phantom,true,imgstd] (x0 tilde next) {$\tilde{x}_0$}
%% node[next,true,action] (a0 next) {$a_0$}
node[next,ztoken,computed] (z1 1 next) {$\hat{z}^1_1$}
node[next,ztoken,computed] (z2 1 next) {$\hat{z}^2_1$}
node[next,empty] (z etc 1 next) {\large $\dots$}
node[next,ztoken,computed] (zKm1 1 next) {$\hat{z}^{K-1}_1$}
node[next,ztoken,computed] (zK 1 next) {$\hat{z}^K_1$}
%% node[next,true,imgstd] (x1 tilde next) {$\tilde{x}_1$}
%% node[next,true,action] (a1 next) {$a_1$}
node[next,empty] (x1 tilde next) {}
node[next,empty] (a1 next) {}
node[next,ztoken,computed] (z1 2 next) {$\hat{z}^1_2$}
node[next,ztoken,computed] (z2 2 next) {$\hat{z}^2_2$}
node[next,empty] (z etc 2 next) {\large $\dots$}
node[next,ztoken,computed] (zKm1 2 next) {$\hat{z}^{K-1}_2$}
node[next,ztoken,computed] (zK 2 next) {$\hat{z}^K_2$}
node[next,empty] (x2 tilde next) {}
node[next,empty] (a2 next) {}
node[next,ztoken,computed] (z1 3 next) {$\hat{z}^1_3$}
node[next,ztoken,computed] (z2 3 next) {$\hat{z}^2_3$}
node[next,empty] (z etc 3 next) {\large $\dots$}
node[next,ztoken,computed] (zKm1 3 next) {$\hat{z}^{K-1}_3$}
node[next,ztoken,computed] (zK 3 next) {$\hat{z}^K_3$}
;

\node[computed,imgstd,above=2pt of {z1 1 next}] (r0 hat) {$\hat{r}_0$};
\node[computed,imgstd,above=2pt of {r0 hat}] (d0 hat) {$\hat{d}_0$};

\node[computed,imgstd,above=2pt of {z1 2 next}] (r1 hat) {$\hat{r}_1$};
\node[computed,imgstd,above=2pt of {r1 hat}] (d1 hat) {$\hat{d}_1$};

\node[computed,imgstd,above=2pt of {z1 3 next}] (r2 hat) {$\hat{r}_2$};
\node[computed,imgstd,above=2pt of {r2 hat}] (d2 hat) {$\hat{d}_2$};

\node[true,imgstd,below=5mm of {x0 tilde}] (x0) {$x_0$};

%% \node[true,imgstd,below=5mm of {x1 tilde},pattern=north east lines,pattern color=red!25] (x1) {$x_1$};
%% \node[true,imgstd,below=5mm of {x2 tilde},pattern=north east lines,pattern color=red!25] (x2) {$x_2$};

\node[fill=blue!35,imgstd,below=5mm of {x1 tilde},postaction={pattern={Lines[angle=-45, distance=4pt,line width=2pt]}, pattern color=red!35,draw}] (x1) {$x_1$};
\node[fill=blue!35,imgstd,below=5mm of {x2 tilde},postaction={pattern={Lines[angle=-45, distance=4pt,line width=2pt]}, pattern color=red!35,draw}] (x2) {$x_2$};

\draw[flow] (x0) -- (x0 tilde);
\draw[flow] (x1) -- (x1 tilde);
\draw[flow] (x2) -- (x2 tilde);

\draw[flow] (a0) -- (z1 1 next);
\draw[flow] (z1 1) -- (z2 1 next);
\draw[flow] (zKm1 1) -- (zK 1 next);

\draw[flow] (a1) -- (z1 2 next);
\draw[flow] (z1 2) -- (z2 2 next);
\draw[flow] (zKm1 2) -- (zK 2 next);

\draw[flow] (a2) -- (z1 3 next);
\draw[flow] (z1 3) -- (z2 3 next);
\draw[flow] (zKm1 3) -- (zK 3 next);

\end{tikzpicture}

\caption{Unrolling dynamics over time.
At each time step (separated by dashed lines), the GPT-like autoregressive transformer $G$ predicts the \dtoken-tokens for the next frame, as well as the reward and a potential episode termination. Its input sequence consists of action tokens, \dtoken-tokens, and \itoken-tokens, namely continuous image embeddings that alleviate the need to attend to past \dtoken-tokens for world modelling. More specifically, an initial frame $x_0$ is embedded into \itoken-token $\tilde{x_0}$. From $\tilde{x_0}$ and  $a_0$, $G$ predicts the reward $\hat{r}_0$, episode termination $\hat{d}_0 \in \{0, 1\}$, and in an autoregressive manner $\hat{z}_1 = (\hat{z}_1^1, \dots, \hat{z}_1^K)$, the \dtoken-tokens for the next frame. Note that, during the imagination procedure, the next frame (stripped box) is computed by the decoder $D$ based on previous frames, actions, and the \dtoken-tokens generated by $G$, i.e. $x_1 = D(x_0, a_0, \hat{z}_1)$.}

\label{fig:autoregressive_model}
\end{figure*}

In the present work, we introduce \diris, a new agent capable of scaling to visually complex environments with lengthier time horizons. \diris\  encodes frames by attending to the ongoing trajectory of observations and actions, effectively describing stochastic deltas between time steps. This enriched conditioning scheme drastically reduces the number of tokens to encode frames, offloads the deterministic aspects of world modelling to the autoencoder, and lets the autoregressive transformer focus on stochastic dynamics. Nonetheless, substituting the sequence of absolute image tokens with a sequence of \dtoken-tokens makes the task of the autoregressive model more arduous. In order to predict the next transition, it may only reason over previous \dtoken-tokens, and thus faces the challenge of integrating over multiple time steps as a way to form a representation of the current state of the world. To resolve this issue, we modify the sequence of the autoregressive model by interleaving continuous \itoken-tokens, that summarize successive world states with frame embeddings, and discrete \dtoken-tokens.

In the Crafter benchmark \cite{crafter}, \diris\ exhibits favorable scaling properties: the agent solves 17 out of 22 tasks after 10M frames of data collection, supersedes DreamerV3 \cite{dreamerv3} at multiple frame budgets, and trains 10 times faster than \textsc{iris}. In addition, we include results in the sample-efficient setting with Atari games. Through experiments, we provide evidence that \diris\ learns to disentangle the deterministic and stochastic aspects of world modelling. Moreover, we conduct ablations to validate the new conditioning schemes for the autoencoder and transformer models.
\section{Method}
\label{sec:method}

We consider a Partially Observable Markov Decision Process (\textsc{pomdp}) \cite{sutton}. The transition, reward, and episode termination dynamics are captured by the conditional distributions $p(x_{t+1} \mid x_{\le t}, a_{\le t})$ and $p(r_t, d_t \mid x_{\le t}, a_{\le t})$, where $x_t \in \mathcal{X} = \mathbb{R}^{3 \times h \times w}$ is an image observation, $a_t \in \mathcal{A} = \{1, \dots, A\}$ a discrete action, $r_t \in \mathbb{R}$ a scalar reward, and $d_t \in \{0, 1\}$ indicates episode termination. The reinforcement learning objective is to find a policy $p_{\pi}(a_{t} \mid x_{\le t}, a_{< t})$ that maximizes the expected sum of rewards $\mathbb{E}_\pi[\sum_{t \ge 0} \gamma^t r_t]$, with discount factor $\gamma \in (0, 1)$.

Learning in imagination \cite{sutton1991dyna, sutton} consists of 3 stages that are repeated alternatively: experience collection, world model learning, and policy improvement. Strikingly, the agent learns behaviours purely within its world model, and real experience is only leveraged to learn the environment dynamics.

In the vein of \textsc{iris} \cite{iris}, our world model is composed of a discrete autoencoder \cite{vqvae} and an autoregressive transformer \cite{vaswani, radford_gpt2}, albeit with new conditioning schemes and architectures. We first expose \textsc{iris}' world model in Section \ref{sec:method:background}, then present \diris' autoencoder and autoregressive model in Sections \ref{sec:method:autoencoder} and \ref{sec:method:dynamics-model}, respectively. Finally, we describe the policy improvement phase in Section \ref{sec:method:policy}. Appendix \ref{app:architectures_hyperparameters} gives a detailed breakdown of model architectures and hyperparameters.

\subsection{Background: \textsc{iris}}
\label{sec:method:background}

High-dimensional images are converted into tokens with a discrete autoencoder $(E_I, D_I)$ \cite{vqvae}. The encoder $E_I : \mathbb{R}^{h \times w \times 3} \rightarrow \{ 1, \dots, N_I \}^{K_I}$ maps an input image $x_t$ into $K_I$ tokens from a vocabulary of size $N_I$. The discretization is done by picking the index of the vector in the vocabulary embedding table that is closest to the encoder output $y_t \in \mathbb{R}^{K_I \times d}$. The $K_I$ tokens are then decoded back into an image with $D_I : \{1, \dots, N_I\}^{K_I} \rightarrow \mathbb{R}^{h \times w \times 3}$. This discrete autoencoder is trained with $L_1$ reconstruction, perceptual \cite{vqgan} and commitment losses \cite{vqvae} computed on collected frames.

The transformer $G_I$ models the environment dynamics by operating over an input sequence of image and action tokens $(z_0^1, \dots, z_0^{K_I}, a_0, z_1^1, \dots, z_1^{K_I}, a_1, \dots, z_t^1, \dots, z_t^{K_I}, a_t)$. Image and action tokens are embedded with learnt lookup tables. At each time step, $G_I$ predicts the transition, reward, and termination distributions: $p_{G_I}(\hat{z}_{t+1} | z_{\le t}, a_{\le t}) \text{ with } \hat{z}_{t+1}^k \, {\sim} \, p_{G_I}(\hat{z}_{t+1}^k | z_{\le t}, a_{\le t}, z_{t+1}^{<k})$, 
 $p_{G_I}(\hat{r}_t | z_{\le t}, a_{\le t})$, and $p_{G_I}(\hat{d}_t | z_{\le t}, a_{\le t})$. The model is trained with a cross-entropy loss on segments sampled from past experience.

 At a high level, the autoencoder builds a vocabulary of image tokens to encode each frame, and the transformer captures the environment dynamics by autoregressively composing the vocabulary over time. As a result, this world model is capable of attending to previous time steps to make its predictions, and models the joint law of future latent states.

\subsection{Disentangling deterministic and stochastic dynamics}
\label{sec:method:autoencoder}

\textsc{iris} \cite{iris} encodes frames independently, making no assumption about temporal redundancy within trajectories. One major drawback of this general formulation is that, in environments with visually challenging frames, a large number of tokens is required to encode frames losslessly. Consequently, computations with the dynamics model become increasingly prohibitive, as the attention mechanism scales quadratically with sequence length. Therefore, limiting computation under such a trade-off may result in degraded performance (\citet{iris} app. E)

\newpage

One possible solution to achieve fast world modelling with minimal loss is to condition the autoencoder on previous frames and actions. Intuitively, encoding a frame given previous frames consists in describing what has changed, the delta, between successive time steps. In many environments, the delta between frames is often much simpler to describe than the frames themselves. As a matter of fact, when the transition function is deterministic, adding previous actions to the conditioning of the decoder results in a world model, without the need to encode any information between time steps. However, most environments of interest feature stochastic dynamics, and apart from aleatoric uncertainty, architectural limitations such as the agent's memory may induce additional epistemic uncertainty. Hence, the delta between two time steps usually consists of deterministic and stochastic components. 

For instance, an agent moving from one square to another in a grid-like environment when pressing movement keys can be seen as a deterministic component of the transition. On the other hand, the sudden apparition of an enemy in a nearby square is a random event. Interestingly, only the stochastic features of a transition should be encoded, and the autoencoder could directly learn to model the deterministic dynamics, which do not require the expressivity and ability to handle multimodality of an autoregressive model. Therefore, when autoenconding frames by conditioning on previous frames and actions, a frame encoding may only consist of a handful of \dtoken-tokens, instead of a large number of image tokens describing frames independently.

Section \ref{sec:experiments:disentanglement} provides empirical evidence that \diris' autoencoder learns to encode frames in such fashion, and Figure \ref{fig:autoencoder} illustrates the new conditioning scheme of the autoencoder.

\begin{figure*}[t]
\begin{center}
\dtoken-tokens sampled randomly \\
\vspace{-0.02cm}
\includegraphics[scale=0.31]{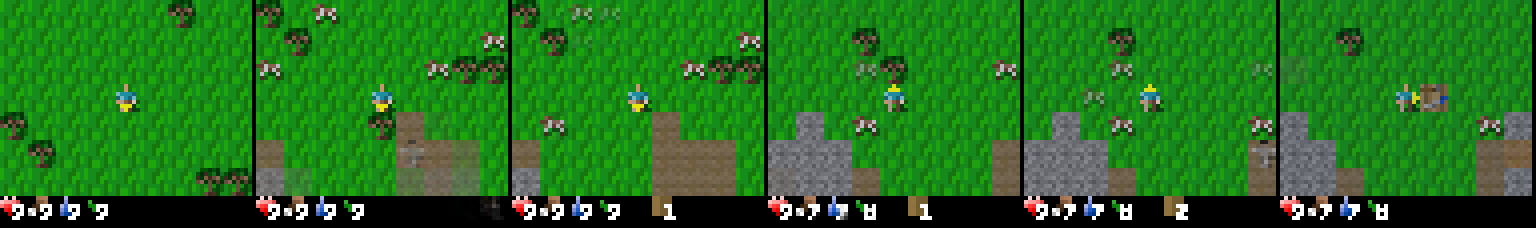} \\
\vspace{-0.05cm}
\dtoken-tokens sampled by the autoregressive transformer \\
\vspace{-0.02cm}
\includegraphics[scale=0.31]{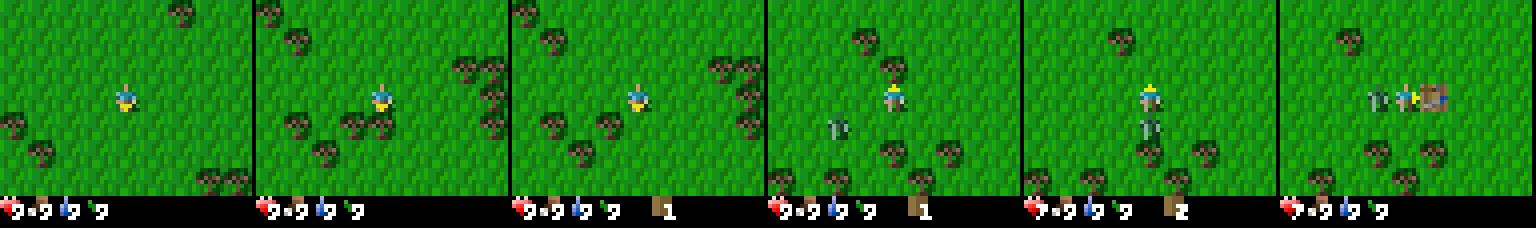} \\
\end{center}
\vspace{-0.33cm}
\hspace{1.12cm} $t = 0$ \hspace{1.85cm} $t = 4$ \hspace{1.84cm} $t = 5$ \hspace{1.86cm} $t = 9$ \hspace{1.84cm} $t = 10$ \hspace{1.68cm} $t = 12$

\caption{Evidence of dynamics disentanglement. Two trajectories are imagined with different ways of generating \dtoken-tokens. In the top trajectory, \dtoken-tokens are sampled randomly. In the bottom trajectory, the autoregressive transformer predicts future \dtoken-tokens. The same starting frame ($t=0$) and sequence of actions are used. With random \dtoken-tokens, the deterministic aspects of the dynamics (layout, movement, items, crafting) are still properly modelled, but the stochastic dynamics (mobs, health indicators) become problematic. For instance, the agent successfully cuts down a tree between $t=4$ and $t=5$, and uses wood planks to build a crafting table between $t=10$ and $t=12$. We observe that these dynamics are modelled in the same way whether \dtoken-tokens are sampled randomly or not. However, in the top trajectory, large quantities of cows appear and disappear from the screen incoherently, whereas the bottom trajectory does not display such erratic patterns. This experiment shows that \diris\ encodes stochastic deltas between time steps with \dtoken-tokens, and its decoder handles the deterministic aspects of world modelling. Appendix \ref{app:fig:disentanglement} contains additional examples.}

\label{fig:disentanglement}
\end{figure*}

More formally, for any set $\mathcal{Y}$, we denote $\mathcal{S}_n (\mathcal{Y}) = \bigcup_{i = 1}^n \mathcal{Y} ^ i$ the set of tuples of elements from $\mathcal{Y}$ of maximum length $n$, and $\mathcal{S} (\mathcal{Y}) = \mathcal{S}_\infty(\mathcal{Y})$. Let $\mathcal{Z} = \{1, \dots, N\}$ a vocabulary of discrete tokens.  Given past images and actions $(x_0, a_0, \dots, x_{t-1}, a_{t-1})$, the encoder $E : \mathcal{S} ( \mathcal{X} \times \mathcal{A} ) \times \mathcal{X} \rightarrow \mathcal{Z} ^ K$ converts an image $x_t$ into $z_t = (z_t^1, \dots, z_t^K)$, a sequence of $K$ discrete  \dtoken-tokens. The encoder is parameterized by a Convolutional Neural Network (\textsc{cnn}) \cite{cnn_lecun}. Actions are embedded with a learnt lookup table and concatenated channel-wise with frames. We use vector quantization \cite{vqvae,vqgan} with factorized and normalized codes \cite{vitvqgan} to discretize the encoder's continuous outputs. The \textsc{cnn} decoder $D : \mathcal{S} ( \mathcal{X} \times \mathcal{A} ) \times \mathcal{Z} ^ K \rightarrow \mathcal{X}$ reconstructs an image $\hat{x}_t$ from past frames, actions and \dtoken-tokens $(x_0, a_0, \dots, x_{t - 1}, a_{t - 1}, z_t)$. Action and \dtoken-tokens are embedded with learnt lookup tables, and concatenated channel-wise with feature maps obtained by forwarding frames through an auxiliary \textsc{cnn}.

The discrete autoencoder is trained on previously collected trajectories with a weighted combination of $L_1$, $L_2$ and max-pixel \cite{anand2022procedural} reconstruction losses, as well as a commitment loss \cite{vqvae}. The codebook is updated with an exponential moving average \cite{vqvae2} and we use a straight-through estimator \cite{straightthrough} to enable backpropagation.

\subsection{Modelling stochastic dynamics}
\label{sec:method:dynamics-model}

While it should be possible to predict future \dtoken-tokens, given a starting image, past actions and \dtoken-tokens, we found this task much more difficult than simply predicting future image tokens, given past image tokens and actions, as in \textsc{iris}. 

To better understand why this is the case, let us consider another example: in a grid environment, \dtoken-tokens may describe the unpredictable movement of an enemy, randomly jumping from one square to another at every time step. Based on the initial enemy location and after only a few time steps, it becomes increasingly difficult to predict if the enemy and the agent are located on the same square, which could trigger a battle and make the enemy disappear. Indeed, situating the two entities involves reasoning about the initial observation, and integrating over all of the previous action and \dtoken-tokens, which may have a complex dependence structure.

To address this problem, we alter the sequence of the dynamics model by interleaving continuous \itoken-tokens, in reference to MPEG's \itoken-frames \cite{richardson2004mpeg}, and discrete \dtoken-tokens. \itoken-tokens alleviate the need of integrating over past \dtoken-tokens to form a representation of the current state of the world, i.e. they deploy a ``soft'' Markov blanket for the prediction of the next \dtoken-tokens.

\begin{table*}[t]
\caption{Returns, number of parameters, and frames collected per second (FPS) for the methods considered. We compute FPS as the total number of environment frames collected divided by the training duration. \diris\ outperforms DreamerV3 for larger frame budgets, and is 10x faster than \textsc{iris} (64 tokens).}
\label{tab:metrics}

\begin{center}
\begin{tabular}{lccccc}
\toprule
Method                     & Return @1M          & Return @5M           & Return @10M          & \#Parameters   & FPS     \\
\midrule
\diris                     & 7.7 (0.5)          & \textbf{15.4} (0.4) & \textbf{16.1} (0.1) & \textbf{25M} & 20          \\
DreamerV3 XL               & \textbf{9.2} (0.3) & 14.2 (0.2)          & 15.1 (0.3)          & 200M         & \textbf{30} \\
\textsc{iris} (64 tokens)  & 5.5 (0.7)          & -                    & -                    & 48M          & 2           \\
\midrule
\diris\ w/o \itoken-tokens & 6.6 (0.2)          & 10.4 (0.5)          & 12.6 (0.8)          & 24M          & 22          \\
DreamerV3 M                & 6.2 (0.5)          & 12.6 (0.7)          & 13.7 (0.8)          & 37M          & 40          \\
\textsc{iris} (16 tokens)  & 4.4 (0.1)          & -                    & -                    & 50M          & 6           \\
\bottomrule
\end{tabular}
\end{center}
\vspace{-0.02\linewidth}
\end{table*}
\begin{figure}[t]
\begin{center}
\includegraphics[scale=0.84]{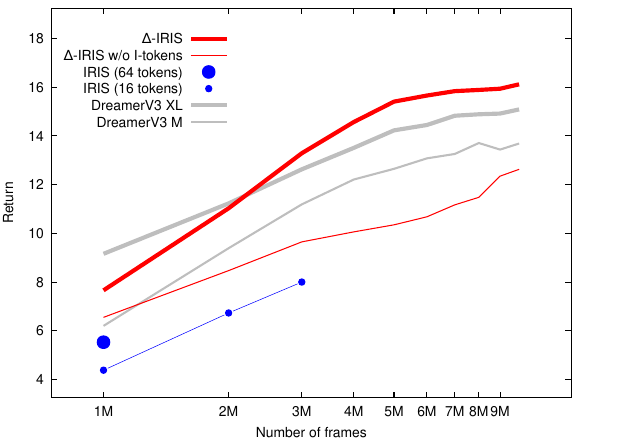}
\end{center}
\vspace{-0.05\linewidth}
\caption{Returns at multiple frame budgets in the Crafter benchmark. \diris\ achieves higher returns than DreamerV3 beyond 3M frames, and surpasses \textsc{iris} for all frame budgets considered. Removing \itoken-tokens from the input sequence of the autoregressive transformer significantly hurts performance.}
\label{fig:results}
\vspace{-0.02\linewidth}
\end{figure}
\begin{figure*}[t]
\begin{center}
\includegraphics[scale=1.15]{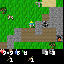}\hspace{0.05cm}
\includegraphics[scale=1.15]{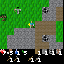}\hspace{0.05cm}
\includegraphics[scale=1.15]{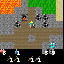}
\hspace{0.4cm}
\includegraphics[scale=1.15]{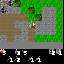}\hspace{0.05cm}
\includegraphics[scale=1.15]{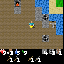}\hspace{0.05cm}
\includegraphics[scale=1.15]{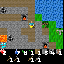}\\

\includegraphics[scale=1.15]{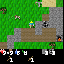}\hspace{0.05cm}
\includegraphics[scale=1.15]{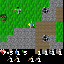}\hspace{0.05cm}
\includegraphics[scale=1.15]{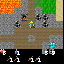}
\hspace{0.4cm}
\includegraphics[scale=1.15]{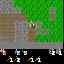}\hspace{0.05cm}
\includegraphics[scale=1.15]{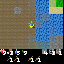}\hspace{0.05cm}
\includegraphics[scale=1.15]{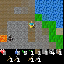}\\

\frame{\includegraphics[scale=1.15]{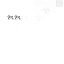}}\hspace{0.05cm}
\frame{\includegraphics[scale=1.15]{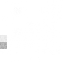}}\hspace{0.05cm}
\frame{\includegraphics[scale=1.15]{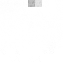}}
\hspace{0.4cm}
\frame{\includegraphics[scale=1.15]{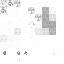}}\hspace{0.05cm}
\frame{\includegraphics[scale=1.15]{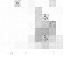}}\hspace{0.05cm}
\frame{\includegraphics[scale=1.15]{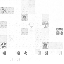}}\\
\end{center}
\hspace{3.0cm} \diris\ 4 tokens \hspace{6.3cm} \textsc{iris} 16 tokens\\
\vspace{-0.5cm}

\caption{Bottom $1\%$ test frames autoencoded by \diris\ (4 tokens) and \textsc{iris} \cite{iris} (16 tokens). Each token takes a value in $\{1, 2, \dots, 1023, 1024\}$, i.e. \diris\ encodes frames with $4 \times \log_{2}(1024) = 40$ bits while \textsc{iris} uses 160 bits. Original frames, reconstructions, and errors are respectively displayed in the top, middle, and bottom rows. Even in the worst instances, \diris\ makes only minor errors, whereas \textsc{iris} fails to accurately reconstruct frames. These errors severely hamper the agent's performance, as it purely learns behaviours from frames generated by its autoencoder.}
\label{fig:reconstructions}
\end{figure*}
\begin{figure*}[t]
\begin{center}
Autoregressive transformer with \itoken-tokens \\
\vspace{-0.02cm}
\includegraphics[scale=0.19]{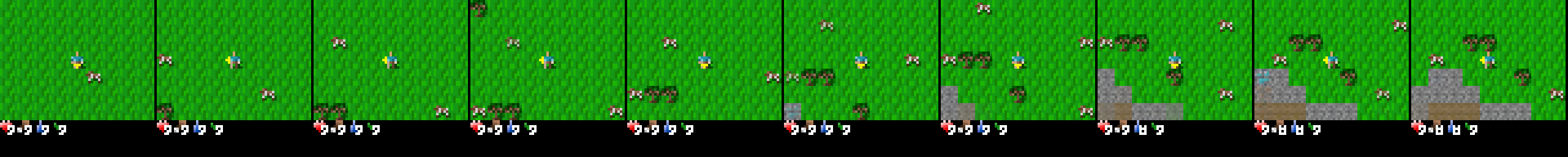} \\
\includegraphics[scale=0.19]{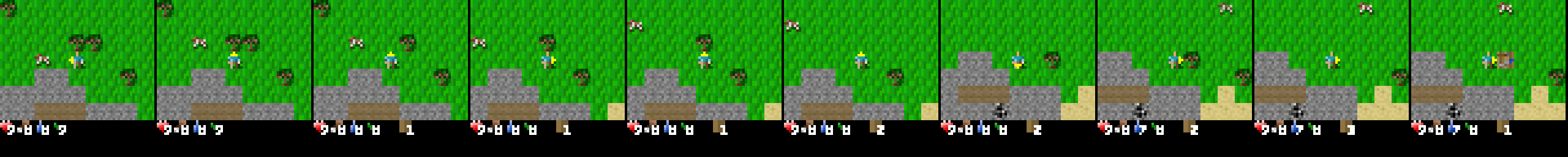} \\
\includegraphics[scale=0.19]{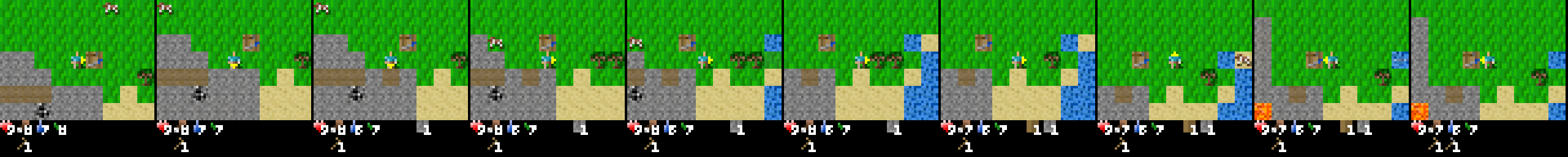} \\
\vspace{-0.05cm}
Autoregressive transformer without \itoken-tokens \\
\vspace{-0.02cm}
\includegraphics[scale=0.25]{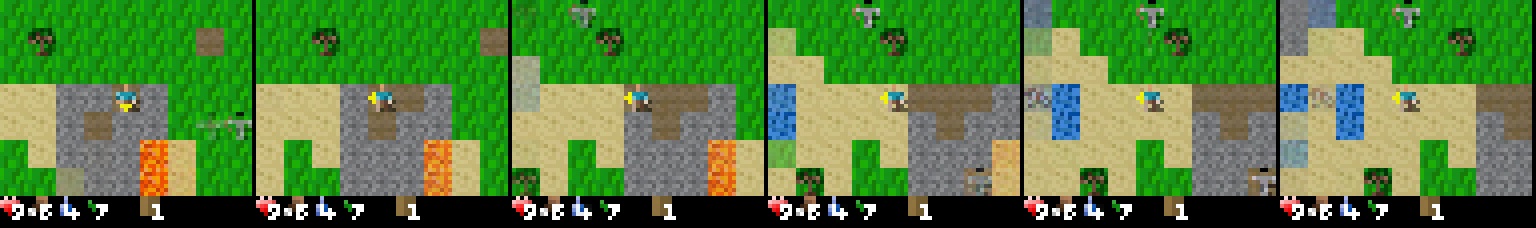} \\
\includegraphics[scale=0.25]{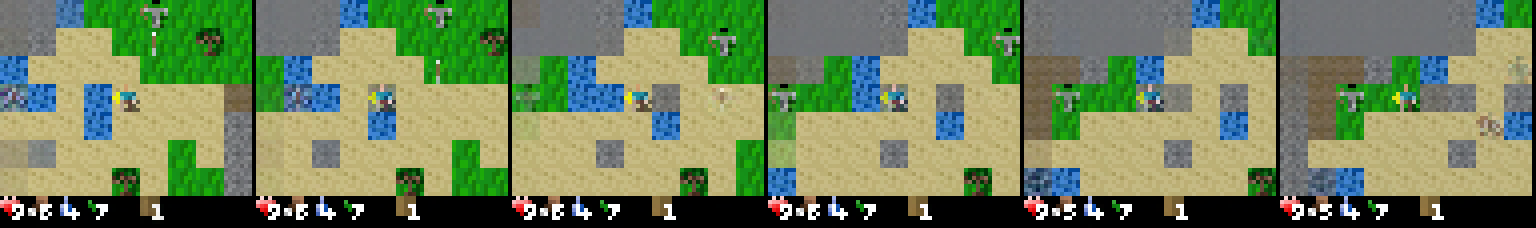} \\
\end{center}
\caption{Trajectories imagined with (top) and without (bottom) \itoken-tokens. In the top trajectory, we observe more than 30 seconds of gameplay generated by \diris' world model. A wide variety of mechanics have been internalized: scrolling, chopping down trees, building a crafting table, mining iron, crafting pickaxes, etc. However, removing \itoken-tokens from the sequence of the autoregressive transformer makes the task of predicting future \dtoken-tokens drastically harder as evidenced by the agent glitching through walls and water in the bottom trajectory. These mistakes ultimately hinder the policy improvement phase, since the agent will reinforce behaviours in a world that does not properly reflect its environment.}
\label{fig:imagination}
\vspace{-0.007\linewidth}
\end{figure*}

\newpage

We obtain \itoken-tokens by forwarding frames through an auxiliary \textsc{cnn} at each time step. They are not produced by a discrete autoencoder. Since \itoken-tokens are not predicted by the model but rather enrich its conditioning, there are no incentives to include a lossy discretization operator or to optimize a reconstruction loss. Instead, they are optimized end-to-end with the learning objectives of the dynamics model. With this improved conditioning, the dynamics model perceives the ongoing trajectory with a mixture of continuous and discrete representations, while making its predictions autoregressively in a discrete space.

Figure \ref{fig:autoregressive_model} displays the input sequence of the dynamics model and the quantities it predicts. Given a sequence of past \itoken-tokens, action tokens, and \dtoken-tokens $(\tilde{x}_0, a_0, z_1^1, \dots, z_1^K, \dots, \tilde{x}_{t-1}, a_{t-1}, z_t^1, \dots, z_t^k)$, the dynamics model $G$ outputs a categorical distribution on $\mathcal{Z}$ for the next \dtoken-token $\hat{z}_t^{k+1} \, {\sim} \, p_{G}(\hat{z}_{t}^{k+1} | \tilde{x}_{< t},  z_{< t}, a_{< t}, z_{t}^{\le k})$. It also predicts distributions for rewards $p_{G}(\hat{r}_{t} | \tilde{x}_{\le t} z_{\le t}, a_{\le t})$ and episode terminations $p_{G}(\hat{d}_{t} | \tilde{x}_{\le t},  z_{\le t}, a_{\le t})$.

$G$ is parameterized by a stack of transformer encoder layers with causal self-attention \cite{vaswani, radford_gpt2}. It is trained with a cross-entropy loss for transition and termination predictions, and we follow DreamerV3 \cite{dreamerv3} in using discrete regression with two-hot targets and symlog scaling for reward prediction \cite{imani2018improving}.

\subsection{Policy improvement}
\label{sec:method:policy}

During the policy improvement phase, the policy $\pi$ learns in the imagination \textsc{pomdp} of its world model, composed of the autoencoder $(E,D)$ and the dynamics model $G$.

At time step $t$, the policy observes a reconstructed image observation $\hat{x}_t$ and samples action $a_t \sim \pi(a_t | \hat{x}_{\le t})$. The world model then predicts the reward $\hat{r}_t$, the episode end $\hat{d}_t$, and the next observation $\hat{x}_{t+1} = D(\hat{x}_{\le t}, \hat{a}_{\le t}, \hat{z}_{\le t}, \hat{z}_{t+1})$, with $\hat{z}_{t+1} \sim p_G(\hat{z}_{t+1} | \hat{x}_{\le t}, \hat{a}_{\le t}, \hat{z}_{\le t})$. The imagination procedure is initialized with a real observation $x_0$ sampled from past experience, and is rolled out for $H$ steps. The procedure stops if an episode termination is predicted before reaching the imagination horizon.

We employ to a large extent the actor-critic training method used for \textsc{iris} \cite{iris}. A value baseline is trained to predict $\lambda$-returns \cite{sutton} with the same discrete regression objective as for reward prediction. The policy optimizes the \textsc{reinforce} with value baseline \cite{sutton} learning objective over imagined trajectories. Exploration is encouraged by adding an entropy maximization term to the policy's objective.
\section{Experiments}
\label{sec:experiments}

In our experiments, we consider the Crafter benchmark \cite{crafter} to illustrate \diris' ability to scale to a visually rich environment with large frame budgets. Besides, we also include Atari 100k games \cite{bellemare13arcade, simple} in Appendix \ref{app:atari_results} to showcase the performance and speed of our agent in the sample-efficient setting.

We introduce the Crafter benchmark and baselines in Section \ref{sec:experiments:benchmark-baselines}. Then, we present our results in Section \ref{sec:experiments:results}. Finally, in Sections \ref{sec:experiments:wm-analysis} and \ref{sec:experiments:disentanglement}, we propose qualitative experiments to validate \diris' world model architecture, and better our understanding of how the model represents information.

\subsection{Benchmark and baselines}
\label{sec:experiments:benchmark-baselines}

Crafter \cite{crafter} is a procedurally generated environment, inspired by the video game Minecraft, with visual inputs, a discrete action space and non-deterministic dynamics. By incorporating mechanics from survival games and a technology tree, this benchmark evaluates a broad range of agent capabilities such as generalization, exploration, and credit assignment. During each episode, the agent's goal is to solve as many tasks as possible, e.g. slaying mobs, crafting items, and managing health indicators. 

Regarding baselines, we consider two model-based RL agents learning in imagination: \textsc{iris} \cite{iris} and DreamerV3 \cite{dreamerv3}. We run several variants: \textsc{iris} (16 tokens), encoding frames with $K_I = 16$ tokens, \textsc{iris} (64 tokens), encoding frames with $K_I = 64$ tokens, and configurations of DreamerV3 of different sizes, namely DreamerV3 XL and DreamerV3 M. To demonstrate the importance of \itoken-tokens, we also run \diris\ without \itoken-tokens in the sequence of the transformer, i.e. $G$ only operates over the first frame as well as actions and \dtoken-tokens.

We keep a fixed imagined-to-collected data ratio of 64 to balance speed and performance. Our experiments run on a Nvidia A100 40GB GPU, with 5 seeds for all methods and ablations. We evaluate each run by computing the average return over 256 test episodes every 1M frames. Note that we stop the \textsc{iris} experiments before 10M frames because they are prohibitively slow.

\subsection{Results}
\label{sec:experiments:results}

Table \ref{tab:metrics} exhibits key metrics and Figure \ref{fig:results} displays learning curves. After 10M frames of data collection, \diris\ solves on average 17 out of 22 tasks, setting a new state of the art for the Crafter benchmark. Beyond the 3M frames mark, \diris\ consistently achieves higher returns than DreamerV3, although DreamerV3 is better suited for the smallest frame budgets. A key difference between the two methods is that \diris\ does not leverage the representations of its world model for policy learning, which may be especially useful in the scarce data regime. As our main objective is to develop world model architectures that scale to complex environments and larger frame budgets, we leave this exploration to future work. \diris\ outperforms \textsc{iris} for all frame budgets considered, while training an order of magnitude faster. Finally, removing \itoken-tokens from the sequence of the dynamics model drastically hurts performance.

\newpage

We believe that achieving higher returns at the 10M frames cap poses a hard exploration problem. Indeed, three of the missing four tasks require crafting new tools in the presence of a nearby crafting table and furnace. Discovering these tools with a naive exploration strategy is highly unlikely, and we have observed only a few occurrences of those events throughout training runs. 

With too few training samples, the world model is unable to internalize these new mechanics and reflect them during the imagination procedure. We hypothesize that a biased data sampling procedure \cite{kauvar2023curious} could be the key to unlock the missing achievements.

\subsection{World model analysis}
\label{sec:experiments:wm-analysis}

In Section \ref{sec:experiments:results}, we validated our design choices for \diris\ with RL experiments. However, downstream RL performance is an imperfect proxy for the quality of a world model due to many possible confounding factors, e.g. the choice of the RL algorithm, entangled world model and policy architectures, or the continual learning loop. In this section, we directly focus on the abilities of the world model.

Figure \ref{fig:reconstructions} illustrates the bottom $1\%$ autoencoded test frames with and without conditioning the autoencoder on the ongoing trajectory (i.e. reconstructions with \diris\ vs \textsc{iris}). With as few as 4 tokens per frame, \diris' autoencoder is able to encode frames with minimal loss. On the other hand, without access to previous frames and actions, and even with 16 tokens, \textsc{iris}' autoencoder produces poor reconstructions.

Figure \ref{fig:imagination} displays trajectories imagined to illustrate whether crucial mechanics have been internalized by the world model, when including \itoken-tokens in the sequence of the autoregressive transformer or not. We observe that, with \itoken-tokens, a multitude of game mechanics are well understood, but in the absence of \itoken-tokens the world model is unable to simulate key concepts. Appendix \ref{app:ablations} includes additional quantitative results.

\subsection{Evidence of dynamics disentanglement}
\label{sec:experiments:disentanglement}

In Section \ref{sec:method:autoencoder}, we argued that, by design, \diris' encoder describes stochastic deltas between timesteps with \dtoken-tokens. In the present section, we propose to exhibit this phenomenon. 

We pick a starting frame and a sequence of actions, and predict two different trajectories with the world model. In one case, we sample future \dtoken-tokens randomly. In the other case, \dtoken-tokens are produced by the autoregressive transformer. We consider a scenario where the agent collects wood then builds a crafting table in Figure \ref{fig:disentanglement}. Appendix \ref{app:disentanglement} displays two other scenarios where the agent explores its surroundings, and where it moves down then stands still.

We observe that, even when sampling \dtoken-tokens randomly, the deterministic aspects of the dynamics are properly modelled: grid layout, agent movement, wood level increasing, crafting table appearing, etc. On the other hand, stochastic dynamics become problematic: skeletons and cows appearing and disappearing, food and water indicators decreasing too early, unlikely quantities of enemies and objects, etc. These observations confirm that \diris\ encodes stochastic deltas between time steps with \dtoken-tokens, and its
decoder handles the deterministic aspects of world modelling.
\section{Related Work}
\label{sec:related}

\subsection*{World Models and imagination}

With Dyna, \citet{sutton1991dyna} introduced the idea of learning behaviours in the imagination of a world model. \citet{worldmodels} went beyond the tabular setting and proposed a new world model architecture, composed of a variational autoencoder \cite{vae} and a recurrent network \cite{lstm, Gers2000}, capable of simulating simple visual environments. Following this breakthrough, multiple generations of Dreamer agents \cite{dreamerv1, dreamerv2, dreamerv3} were developed, with DreamerV2 being the first imagination-based agent to outperform humans in Atari games, and DreamerV3 being the first world model architecture applicable to a wide range of domains without any specific tuning. DreamerV2 learns in the imagination of a world model combining a convolutional autoencoder with a recurrent state-space model (RSSM) \cite{planet}. The key modifications that enabled DreamerV2 to improve over the original Dreamer agent were categorical latents and KL balancing between prior and posterior estimates. DreamerV3 builds upon DreamerV2 with more universal design choices such as symlog scaling of rewards and values, combining free bits \cite{freebits} with KL balancing, return scaling for static entropy regularization, and architectural novelties for model scaling. Variants of Dreamer such as TransDreamer \cite{transdreamer} and STORM \cite{zhang2023storm} have also been explored, where transformers replace the recurrent network in the RSSM for dynamics prediction.

A potential limitation of RSSM-like architectures is that they do not model the joint distribution of future latent states, and instead predict product laws. One way to mitigate this discrepancy between the predicted distributions and the distributions of interest is to encourage factorized distributions \cite{dreamerv3}. On the other hand, autoregressive architectures \cite{iris} do model the joint distribution and do not require to enforce independence, which may result in a more expressive model.

\subsection*{Trajectory and video autoencoders}

The idea of encoding frames with respect to past frames predates modern deep learning, and is at the origin of efficient video compression algorithms, such as MPEG \cite{richardson2004mpeg}. In recent years, multiple works have implemented variants of this approach. \citet{vqplanning} propose an offline version of MuZero \citep{muzero} equipped with an autoregressive transformer that performs search over trajectory-level discrete latent variables and actions. Phenaki \cite{phenaki} is a text-to-video model composed of a spatio-temporal discrete autoencoder and a masked bidirectional transformer. TECO \cite{yan2022teco} is an action-conditional video prediction model composed of a discrete frame autoencoder conditioned on the previous frame, a temporal autoregressive transformer, and a spatial MaskGit \cite{chang2022maskgit}. While these methods also encode frames by conditioning on past frames, their dynamics models purely operate over discrete tokens, and do not leverage continuous tokens to alleviate the need to integrate over multiple time steps in order to make the next prediction.

\citet{planet} acknowledge that modelling stochastic dynamics may be difficult, as it would involve remembering information from previous time steps. The authors propose to solve this problem by carrying a ``deterministic'' state over time via a recurrent network, at the core of their RSSM. We make a similar observation, and further show that this task is still difficult even when past information does not have to be carried by a recurrent state, as a transformer can attend to all previous \dtoken-tokens. Hence, it is not only a memory problem, but also a modelling one. Here, we address this issue in a manner that is compatible with autoregressive transformers, namely by injecting continuous \itoken-tokens in the sequence of the dynamics model.

\section{Conclusion}
\label{sec:conclusion}

We introduced \diris, a new model-based agent relying on an efficient world model architecture to simulate its environment and learn new behaviours. \diris\ features a discrete autoencoder that encodes the stochastic aspects of world modelling with discrete \dtoken-tokens, and an autoregressive transformer leveraging continuous \itoken-tokens to model stochastic dynamics.

\newpage

Through experiments, we showed the ability of our agent to scale to the challenging Crafter benchmark, as well as its sample efficiency in Atari100k. Finally, we illustrated how its world model internalized environment dynamics, and conducted ablations to validate our proposed design choices.

In its current form, \diris\  uses the same number of tokens to encode stochastic dynamics at each time step. However, the reality of most environments is such that periods of low uncertainty are quickly followed by moments of high randomness. Therefore, an improved version of the world model could possibly predict dynamically various numbers of tokens based on the current context. Besides, leveraging the internal representations of the world model could potentially result in a lightweight and more robust policy.

\section*{Impact Statement}
The deployment of autonomous agents in real-world applications raises safety concerns. Agents learning new behaviours may harm individuals and damage property. With world models, we lower the amount of time spent interacting with the real world and thus mitigate risks. In this work, we propose a world model architecture that is amenable to scaling up to complex environments, where accurate simulations are even more critical given the usually higher stakes.

\section*{Acknowledgements}
We would like to thank Adam Jelley, Bálint Máté, Daniele Paliotta, Maxim Peter, Youssef Saied, Atul Sinha, and Alessandro Sordoni for insightful discussions and comments. Vincent Micheli was supported by the Swiss National Science Foundation under grant number FNS-187494.

\bibliography{main}
\bibliographystyle{icml2024}

%%%%%%%%%%%%%%%%%%%%%%%%%%%%%%%%%%%%%%%%%%%%%%%%%%%%%%%%%%%%%%%%%%%%%%%%%%%%%%%
%%%%%%%%%%%%%%%%%%%%%%%%%%%%%%%%%%%%%%%%%%%%%%%%%%%%%%%%%%%%%%%%%%%%%%%%%%%%%%%
% APPENDIX
%%%%%%%%%%%%%%%%%%%%%%%%%%%%%%%%%%%%%%%%%%%%%%%%%%%%%%%%%%%%%%%%%%%%%%%%%%%%%%%
%%%%%%%%%%%%%%%%%%%%%%%%%%%%%%%%%%%%%%%%%%%%%%%%%%%%%%%%%%%%%%%%%%%%%%%%%%%%%%%
\newpage
\appendix
\onecolumn

%%%%%%%%%%%%%%%%%%%%%%%%%%%%%%%%%%%%%%%%%%%%%%

\section{Architectures and hyperparameters}
\label{app:architectures_hyperparameters}

\subsection{Discrete autoencoder}

\begin{table}[h]
  \caption{Encoder / Decoder hyperparameters. We list the hyperparameters for the encoder, the same ones apply for the decoder.}
  \centering
  \begin{tabular}{lc}
    \toprule
    Hyperparameter     & Value \\
    \midrule
    Frame dimensions (h, w) & $64 \times 64$ \\
    Layers & 5 \\
    Residual blocks per layer & 2 \\
    Channels in convolutions per layer & $[64, 64, 128, 128, 256]$ \\
    Downsampling after layer $n$ & $[1, 0, 1, 1, 0]$ \\
    \midrule
    Past actions embedding channels & 4 \\
    Decoder past frames embedder arch. & Same as encoder \\
    Decoder past frames embedder output feature map size & $8 \times 8 \times 8$ \\
    Conditioning time steps & $1$ \\
    \midrule
    $L_1$ loss weight & 0.1 \\
    $L_2$ loss weight & 1.0 \\
    Max-pixel loss weight & 0.01 \\
    Commitment loss weight & 0.02 \\
    \bottomrule
  \end{tabular}
\end{table}

\begin{table}[h]
  \caption{Embedding table and latent state hyperparameters.}
  \centering
  \begin{tabular}{lc}
    \toprule
    Hyperparameter      & Value \\
    \midrule
    Vocabulary size (N) & 1024 \\
    Tokens per frame (K)& 4  \\
    Latent feature map size & $64 \times 8 \times 8$ \\
    Pre-discretization token size & $64 \times 4 \times 4$ \\
    Token embedding dimension & 64 \\
    Codebook moving average coefficient & 0.99 \\
    \bottomrule
  \end{tabular}
\end{table}

In early experiments, we used a transformer instead of a \textsc{cnn} for the architecture of the autoencoder. It had a much longer context size of twenty time steps. Although the transformer-based autoencoder performed better than its \textsc{cnn} counterpart on static datasets, we observed that the \textsc{cnn} would learn faster than the transformer in the continual learning setup. Besides, for the sake of simplicity, we decreased the initial conditioning of the \textsc{cnn} autoencoder from four time steps to one time step, as the slight increase in reconstruction losses did not significantly hinder agent performance. These observations are largely environment-dependent, thus the context size or the architecture of the autoencoder should most likely be adapted accordingly.

\subsection{Autoregressive transformer}

\begin{table}[h]
  \caption{Transformer hyperparameters.}
  \centering
  \begin{tabular}{lc}
    \toprule
    Hyperparameter     & Value \\
    \midrule
    Timesteps      & 21 \\
    Embedding dimension & 512 \\
    Layers & 3 \\
    Attention heads & 8 \\
    Weight decay & 0.01 \\
    \itoken-token frame embedder arch. & Same as encoder with halved channels per layer \\
    \bottomrule
  \end{tabular}
\end{table}

\subsection{Actor-Critic}

We tie the weights of the actor and critic, except for the last layer. The actor-critic takes as input a frame, and forwards it through a convolutional neural network \cite{cnn_lecun} followed by an LSTM cell \cite{lstm,Gers2000,a3c}. For the \textsc{cnn}, we use the same architecture as the encoder, except that we halve the number of channels per layer. The dimension of the LSTM hidden state is 512.

Before starting the imagination procedure ($H = 15$) from a given frame, we burn-in \cite{r2d2} the 5 previous frames to initialize the hidden state. The discount factor $\gamma$ is 0.997, the parameter for $\lambda$-returns is set to 0.95, and the coefficient for the entropy maximization term is 0.001. Targets for value estimates are produced by a moving average of the critic network, with update parameter 0.995 \cite{dqn}

\subsection{Training loop and shared hyperparameters}

\begin{table}[h]
  \caption{Training loop and shared hyperparameters.}
  \label{tables:shared-hparams}
  \centering
  \begin{tabular}{lc}
    \toprule
    Hyperparameter     & Value \\
    \midrule
    Epochs  & 1000 \\
    Environment steps first epoch & 100000 \\
    Environment steps per epoch & 10000 \\
    \# Collection epochs & 990 \\
    Collection epsilon-greedy & 0.01 \\
    Training steps per epoch & 500 \\
    \bottomrule
  \end{tabular}
  \hspace{2cm}
  \begin{tabular}{lc}
    \toprule
    Hyperparameter     & Value \\
    \midrule
    Autoencoder batch size & 32 \\
    Transformer batch size & 32 \\
    Actor-critic batch size & 86 \\
    Learning rate & 1e-4 \\
    Optimizer & Adam \\
    Max gradient norm & 10.0 \\
    \bottomrule
  \end{tabular}
\end{table}

As mentioned in Section \ref{sec:method}, the world model and policy are trained with temporal segments sampled from past experience. We use a count-based sampling procedure over the entire history of episodes, i.e. the likelihood that a given episode is chosen to produce the next sample is inversely proportional to the number of times it was previously used. We raise inverse counts to the power of 5 to further limit the bias towards older episodes.

%%%%%%%%%%%%%%%%%%%%%%%%%%%%%%%%%%%%%%%%%%%%%%

\section{Impact of design choices on key world modelling metrics}
\label{app:ablations}

Metrics are computed on a held-out test set after training various world models on a dataset consisting of 10M frames collected by a \diris\ agent throughout its training.

\begin{table}[h]
  \caption{Left: Impact of removing past frames and actions from the conditioning of the autoencoder (\diris\ $\rightarrow$ \textsc{iris}). Right: Impact of removing \itoken-tokens from the conditioning of the autoregressive transformer.}
  \label{tables:shared-ablation}
  \centering
  \begin{tabular}{lc}
    \toprule
    Method     & $L_2$ loss \\
    \midrule
    \diris\ (4 tokens)  & 0.000185 \\
    \textsc{iris}~~~~ (64 tokens) & 0.001715 \\
    \textsc{iris}~~~~ (16 tokens) & 0.007496 \\
    \bottomrule
  \end{tabular}
  \hspace{1cm}
  \begin{tabular}{lcc}
    \toprule
    Method     & Next token loss (CE) & Reward loss (CE) \\
    \midrule
    \diris & 1.57 & 0.108 \\
    \diris\ w/o \itoken-tokens & 1.73 & 0.135 \\
    \bottomrule
  \end{tabular}
\end{table}

\begin{table}[h]
  \caption{Impact of discarding the auxiliary max-pixel loss.}
  \label{tables:shared-ablation-losses}
  \centering
  \begin{tabular}{lcc}
    \toprule
    Method     & $L_2$ loss & Max-pixel loss \\
    \midrule
    \diris\  & 0.000185 & 0.018 \\
    \diris\ w/o max-pixel loss & 0.000178 & 0.031 \\
    \bottomrule
  \end{tabular}
\end{table}

\newpage

%%%%%%%%%%%%%%%%%%%%%%%%%%%%%%%%%%%%%%%%%%%%%%

\section{Atari 100k}
\label{app:atari_results}

The Atari 100k benchmark \cite{simple} features Atari games \cite{bellemare13arcade} with diverse mechanics. The specificity of this benchmark is the hard constraint on the number of interactions, namely one hundred thousand per environment. Compared to the standard Atari benchmark, this constraint results in a dramatic drop in real-time experience, from 900 hours to 2 hours.

Regarding baselines, we consider four model-based RL agents learning in imagination: SimPLe \cite{simple}, DreamerV3 \cite{dreamerv3}, \textsc{storm} \cite{zhang2023storm}, and \textsc{iris} \cite{iris}. We note that the current best performing methods for Atari 100k resort to other approaches, such as lookahead search for EfficientZero \cite{efficientzero}, or self-supervised representation learning with periodic resets for BBF \cite{schwarzer2023bigger}.

The usual metric of interest is the HNS, the human-normalized score, based on the performance of human players with similar experience. A negative HNS indicates worse than random performance whereas an HNS above 1 signifies superhuman performance. We evaluate \diris\ by computing an average over 100 episodes collected at the end of training for each game (5 seeds). For the baselines, we report the published results.

Table \ref{app:tab:atari_results} displays returns across games and aggregate metrics \cite{rledge}. \diris\ achieves higher aggregate metrics than \textsc{iris}, while training in 26 hours, a 5-fold speedup.

\begin{table}[h]
    \caption{Returns on the 26 games of Atari 100k after 2 hours of real-time experience, and human-normalized aggregate metrics.}
    \label{app:tab:atari_results}
\begin{center}
\centering
\centering
\begin{tabular}{lrr rrrrr}
\toprule
Game                 &  Random    &  Human    &  SimPLe    &  DreamerV3      &  \textsc{storm}  &  \textsc{iris}    &  \diris\ (ours)     \\
\midrule
Alien                &  228       &  7128     &  617       &  959            &  \textbf{984}    &  420              &  391                \\
Amidar               &  6         &  1720     &  74        &  139            &  \textbf{205}    &  143              &  64                 \\
Assault              &  222       &  742      &  527       &  706            &  801             &  \textbf{1524}    &  1123               \\
Asterix              &  210       &  8503     &  1128      &  932            &  1028            &  854              &  \textbf{2492}                 \\
BankHeist            &  14        &  753      &  34        &  649            &  641             &  53               &  \textbf{1148}                 \\
BattleZone           &  2360      &  37188    &  4031      &  12250          &  \textbf{13540}  &  13074            &  11825               \\
Boxing               &  0         &  12       &  8         &  78             &  \textbf{80}     &  70               &  70                  \\
Breakout             &  2         &  31       &  16        &  31             &  16              &  84               &  \textbf{302}       \\
ChopperCommand       &  811       &  7388     &  979       &  420            &  \textbf{1888}   &  1565             &  1183                 \\
CrazyClimber         &  10781     &  35829    &  62584     &  \textbf{97190} &  66776           &  59324            &  57864                 \\
DemonAttack          &  152       &  1971     &  208       &  303            &  165             &  \textbf{2034}    &  533                 \\
Freeway              &  0         &  30       &  17        &  0              &  \textbf{34}     &  31               &  31        \\
Frostbite            &  65        &  4335     &  237       &  909            &  \textbf{1316}   &  259              &  279                 \\
Gopher               &  258       &  2413     &  597       &  3730           &  \textbf{8240}   &  2236             &  6445     \\
Hero                 &  1027      &  30826    &  2657      &  \textbf{11161} &  11044           &  7037             &  7049                 \\
Jamesbond            &  29        &  303      &  101       &  445            &  \textbf{509}    &  463              &  309                \\
Kangaroo             &  52        &  3035     &  51        &  4098           &  \textbf{4208}   &  838              &  2269                 \\
Krull                &  1598      &  2666     &  2205      &  7782           &  \textbf{8413}   &  6616             &  5978               \\
KungFuMaster         &  259       &  22736    &  14863     &  21420          &  \textbf{26182}  &  21760            &  21534              \\
MsPacman             &  307       &  6952     &  1480      &  1327           &  \textbf{2674}   &  999              &  1067                 \\
Pong                 &  -21       &  15       &  13        &  18             &  11              &  15               &  \textbf{20}        \\
PrivateEye           &  25        &  69571    &  35        &  882            &  \textbf{7781}   &  100              &  103                 \\
Qbert                &  164       &  13455    &  1289      &  3405           &  \textbf{4523}   &  746              &  1444                 \\
RoadRunner           &  12        &  7845     &  5641      &  15565          &  \textbf{17564}  &  9615             &  10414              \\
Seaquest             &  68        &  42055    &  683       &  618            &  525             &  661              &  \textbf{827}                 \\
UpNDown              &  533       &  11693    &  3350      &  \textbf{9234}  &  7985            &  3546             &  4072                 \\
\midrule
\#Superhuman         &  0         &  N/A      &  1         &  9              &  10              &  10               &  \textbf{11}        \\
Mean                 &  0.00      &  1.00     &  0.33      &  1.10           &  1.27            &  1.05             &  \textbf{1.39}      \\
Interquartile Mean   &  0.00      &  1.00     &  0.13      &  0.50           &  0.64            &  0.50             &  \textbf{0.65}               \\
\bottomrule
\end{tabular}
\end{center}
\end{table}

\newpage

%%%%%%%%%%%%%%%%%%%%%%%%%%%%%%%%%%%%%%%%%%%%%%

\section{Crafter scores and individual success rates}

\begin{table}[h]
  \caption{Crafter scores, i.e. geometric mean of success rates.}
  \centering
  \begin{tabular}{lccc}
    \toprule
    Method     & Crafter score @1M & Crafter score @5M & Crafter score @10M \\
    \midrule
    \diris      & 9.30 & 39.67 & 42.47 \\
    \diris\ w/o \itoken-tokens & 5.85 & 14.39 & 25.92 \\
    \textsc{iris} (64 tokens) & 6.66 & - & - \\
    \bottomrule
  \end{tabular}
\end{table}

\begin{figure}[h]
\begin{center}
\includegraphics[scale=0.7]{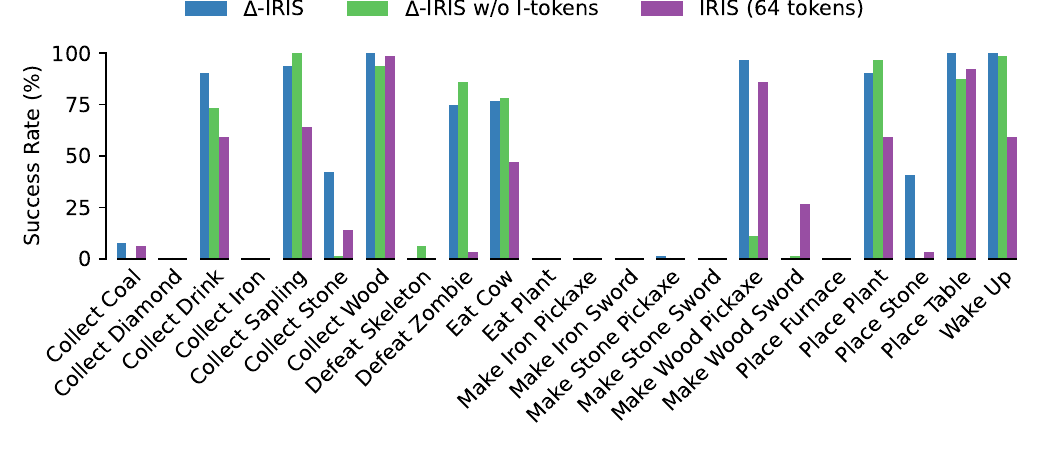}
\end{center}
\caption{Individual success rates after collecting 1M frames.}
\label{app:fig:srates_1}
\end{figure}

\begin{figure}[h]
\begin{center}
\includegraphics[scale=0.7]{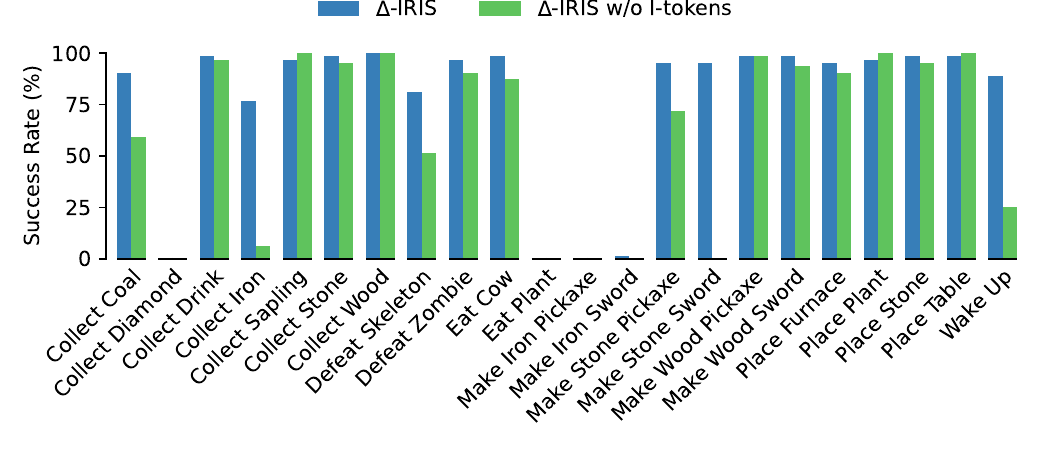}
\end{center}
\caption{Individual success rates after collecting 10M frames.}
\label{app:fig:srates_2}
\end{figure}

%%%%%%%%%%%%%%%%%%%%%%%%%%%%%%%%%%%%%%%%%%%%%%

\section{Baselines}
\label{app:baselines}

DreamerV3 results were obtained with commit \href{https://github.com/danijar/dreamerv3/tree/8fa35f83eee1ce7e10f3dee0b766587d0a713a60}{8fa35f8}. We used the standard configuration for Crafter, and set the \textit{run.train\_ratio} variable controlling the imagined-to-collected data ratio to 64. Note that a new version of DreamerV3 was recently released in April 2024. This update includes additional and broadly applicable novelties for world model and policy learning. 

\textsc{iris} results were obtained with commit \href{https://github.com/eloialonso/iris/tree/ac6be401fed2b6176c9ce0cf1dc10e376c9d740d}{ac6be40}. For the training loop and shared hyperparameters, we picked the same values as in Table \ref{tables:shared-hparams}. We increased the dimension and attention heads of the transformer from 256 and 4 to 512 and 8, respectively. Finally, we used a replay buffer with a capacity of 1M frames.

%%%%%%%%%%%%%%%%%%%%%%%%%%%%%%%%%%%%%%%%%%%%%%

\newpage
\section{Evidence of dynamics disentanglement}

\label{app:disentanglement}
\begin{figure*}[h]
\begin{center}
\dtoken-tokens sampled randomly \\
\vspace{-0.02cm}
\includegraphics[scale=0.31]{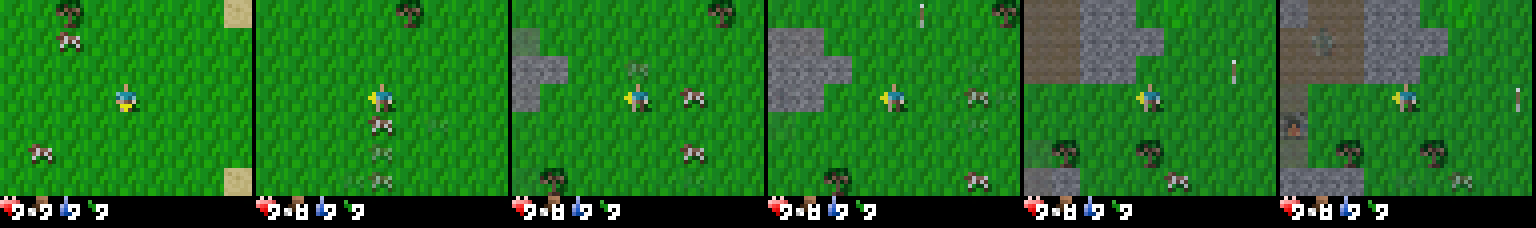} \\
\vspace{-0.05cm}
\dtoken-tokens sampled by the autoregressive transformer \\
\vspace{-0.02cm}
\includegraphics[scale=0.31]{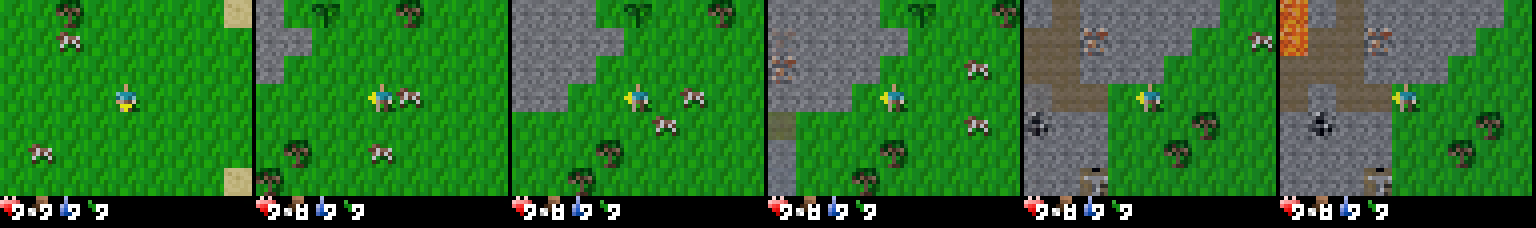} \\
\end{center}
\vspace{-0.3cm}
\hspace{1.12cm} $t = 0$ \hspace{1.85cm} $t = 4$ \hspace{1.84cm} $t = 5$ \hspace{1.86cm} $t = 9$ \hspace{1.84cm} $t = 10$ \hspace{1.68cm} $t = 12$
\vspace{0.5cm}
\begin{center}
\dtoken-tokens sampled randomly \\
\vspace{-0.02cm}
\includegraphics[scale=0.31]{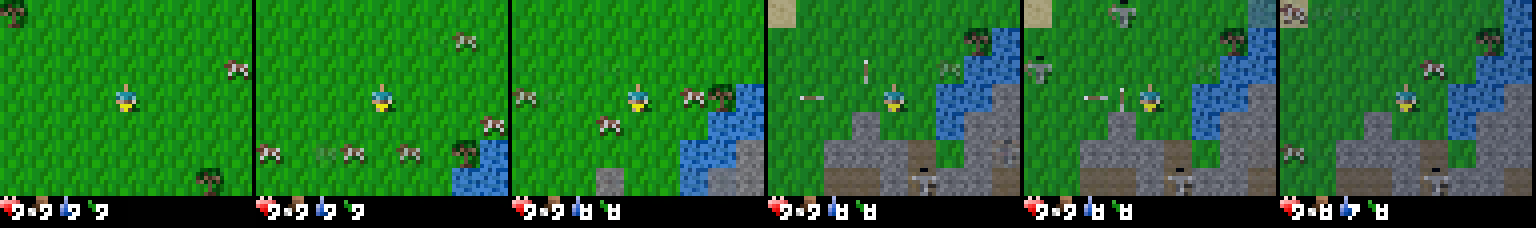} \\
\vspace{-0.05cm}
\dtoken-tokens sampled by the autoregressive transformer \\
\vspace{-0.02cm}
\includegraphics[scale=0.31]{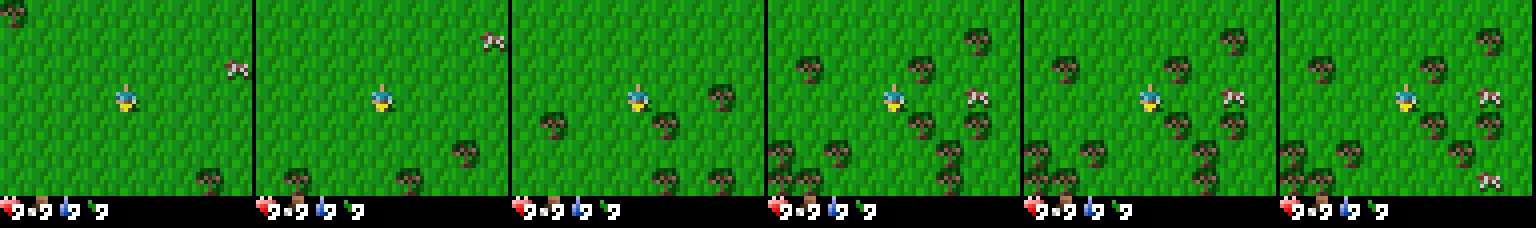} \\
\end{center}
\vspace{-0.3cm}
\hspace{1.12cm} $t = 0$ \hspace{1.85cm} $t = 4$ \hspace{1.84cm} $t = 5$ \hspace{1.86cm} $t = 9$ \hspace{1.84cm} $t = 10$ \hspace{1.68cm} $t = 12$
\label{app:fig:disentanglement}
\caption{Two additional examples of dynamics disentanglement, as discussed in Section \ref{sec:experiments:disentanglement} and Figure \ref{fig:disentanglement}.}
\end{figure*}
\newpage

%%%%%%%%%%%%%%%%%%%%%%%%%%%%%%%%%%%%%%%%%%%%%%

%%%%%%%%%%%%%%%%%%%%%%%%%%%%%%%%%%%%%%%%%%%%%%%%%%%%%%%%%%%%%%%%%%%%%%%%%%%%%%%
%%%%%%%%%%%%%%%%%%%%%%%%%%%%%%%%%%%%%%%%%%%%%%%%%%%%%%%%%%%%%%%%%%%%%%%%%%%%%%%

\end{document}